\documentclass[conference]{IEEEtran}
\IEEEoverridecommandlockouts
\usepackage{cite}
\usepackage{amsmath,amssymb,amsfonts}
\usepackage{algorithmic}
\usepackage{graphicx}
\usepackage{textcomp, ,nicefrac}
\usepackage{babel}
\def\BibTeX{{\rm B\kern-.05em{\sc i\kern-.025em b}\kern-.08em
    T\kern-.1667em\lower.7ex\hbox{E}\kern-.125emX}}
    
\usepackage{float}
\usepackage[table]{xcolor}
\usepackage{collcell}
\usepackage{hhline}
\usepackage{pgf}
\usepackage{multirow}
\usepackage{pbox}

\def\colorModel{rgb} 

\newcommand\ColCell[1]{
\pgfmathparse{#1<0.5?1:0}  
    \ifnum\pgfmathresult=0\relax\color{white}\fi
  \pgfmathsetmacro\compA{1-#1}      
  \pgfmathsetmacro\compB{1-#1} 
  \pgfmathsetmacro\compC{1}      
  \edef\x{\noexpand\centering\noexpand\cellcolor[\colorModel]{\compA,\compB,\compC}}\x #1
  } 
\newcolumntype{E}{>{\collectcell\ColCell}m{0.55cm}<{\endcollectcell}}  
\newcommand*\rot{\rotatebox{90}}

\begin{document}

\title{“Prompt-Gamma Neutron Activation Analysis (PGNAA)” Metal Spectral Classification using Deep Learning Method\\
\thanks{The MetalClass project is funded by the German Federal Ministry of Education and Research. The copper alloys were provided by Wieland-Werke AG. We would also like to express our thanks to the pioneer researchers in deep image classification and other related fields, and to all members who have made a positive contribution to the MetalClass project.}
}

\author{
\IEEEauthorblockN{Ka Yung Cheng}
\IEEEauthorblockA{(Graduate master student at \\
OWL University of Applied Sciences and Arts) \\
\textit{currently with the Institute of Medical Informatics and Statistics} \\
\textit{University Hospital Schleswig-Holstein (UKSH), Campus Kiel}\\
Kaistraße 101, 24114 Kiel, Germany \\
kayung.cheng@uksh.de}
\and
\IEEEauthorblockN{Helmand Shayan}
\IEEEauthorblockA{\textit{Department of Electrical Engineering \& Computer Science} \\
\textit{OWL University of Applied Sciences and Arts}\\
Campusallee 12, 32657 Lemgo, Germany \\
helmand.shayan@th-owl.de}
\and
\IEEEauthorblockN{Kai Krycki}
\IEEEauthorblockA{\textit{Department of Mathematical Methods \& Software Development} \\
\textit{AiNT GmbH}\\
52222 Stolberg, Germany \\
krycki@nuclear-training.de}
\and
\IEEEauthorblockN{Markus Lange-Hegermann}
\IEEEauthorblockA{\textit{Department of Electrical Engineering \& Computer Science} \\
\textit{OWL University of Applied Sciences and Arts}\\
Campusallee 12, 32657 Lemgo, Germany \\
markus.lange-hegermann@th-owl.de}
}

\maketitle

\begin{abstract}
There is a pressing market demand to minimize the test time of Prompt Gamma Neutron Activation Analysis (PGNAA) spectra measurement machine, so that it could function as an instant material analyzer \cite{pgnaa1,pgnaa2,pgnaa3}, e.g. to classify waste samples instantaneously and determine the best recycling method based on the detected compositions of the testing sample.

This article introduces a new development of the deep learning classification and contrive to reduce the test time for PGNAA machine. We propose both Random Sampling Methods and Class Activation Map (CAM) to generate “downsized” samples and train the CNN model continuously. Random Sampling Methods (RSM) aims to reduce the measuring time within a sample, and Class Activation Map (CAM) is for filtering out the less important energy range of the downsized samples.
 
We shorten the overall PGNAA measuring time down to 2.5 seconds while ensuring the accuracy is around 96.88\% for our dataset with 12 different species of substances. Compared with classifying different species of materials, it requires more test time (sample count rate) for substances having the same elements to archive good accuracy. For example, the classification of copper alloys requires nearly 24 seconds test time to reach 98\% accuracy.
\end{abstract}


\section{Introduction}
\label{sec:introduction}
\IEEEPARstart{T}{he} composition of materials can be analyzed in a destructive way using traditional chemical process;  nevertheless, we use an innovative material analysis system, based on PGNAA technology and machine learning classification algorithms. The composition of waste material could now be classified in a non-destructive way and with advanced speed.

Neutron activation analysis (NAA) utilizes neutron-induced reactions to examine the spectra of the emissions of a radioactive sample, which could determine the element concentrations or material compositions of the sample. The sample is firstly bombarded by neutrons, causing the elements to form radioactive isotopes. The new forming activation product can be characterized through their radioactive emissions and the energy spectrum is detected by a sensor. and radioactive decay chain; therefore, metal compositions could be found. This project utilizes PGNAA \cite{pgnaa1}, which is a subcategory of NAA. The additional prefix “PG” means prompt gamma-ray and hinted at state-of-the-art sensors which have a good resolution to detect the emission.

There is a strong marketing demand to apply PGNAA machines to production applications such as recycle factories or ore refineries \cite{pgnaa1,pgnaa2,pgnaa3}. If the necessary measurement time for PGNAA is a few hours, such measurements are only useful in a laboratory setting. The measuring time for production must be “heavily reduced”. However, cutting the measuring time of PGNAA means the output spectrum is incomplete and noisy \cite{noise}. 

To deal with this incompleteness and noise, we devise Random Sampling Methods (RSMs) for generating the “down-sizing” training samples, by assuming the spectral data as probability distributions.
The result spectrum keep the intrinsic down-sized nature while the measurement time of the target sample is drastically cut. 

An additional trick to downsize the samples - visualizing and indicating the importance of each energy level using Class Activation Map (CAM). We can then discard the less important energy range in order to reduce the necessary number of channels in detectors or to allow a finer resolution in the remaining energy ranged. We use these downsized samples to train Convolutional Neural Network (CNN) models.
Since downsizing too much causes negative effects, we closely monitor loss, accuracy, training time, and prediction time throughout the training process. Visualization such as Confusion Matrix was done to ensure the prediction is accurate even the sampling time is minimized. Though other strategies may achieve better result in speed and accuracy \cite{helmand}, CNN is still an attractive tool for general classify purpose due to its interpretability of channel importance. 

\section{Dataset}
We use totally three datasets for our experiments in Section \ref{sec:results}. At the beginning of calibration test, the first dataset, with \textit{10 species of substances (e.g. scrap metal powder, cement and stucco) and two species of soil}, was used for Experiment I. However, we also collected two data sets of samples in the “same species”. The second dataset is only with \textit{aluminium} (for Experiment II and IIIa) and the third is with \textit{copper alloys} from Wieland Electric GmbH (for Experiment IIIb). Each species creates a distinctive PGNAA spectrum. The kinetic energies of electron emission spread over a wide range of energy levels. An individual CSV file is dedicated to each species, which records “Energy (keV)” vs “Counts (intensities)”. The following characteristics can be learned from a PGNAA spectrum:

\begin{enumerate}
\item Each constituent material of a sample could be identified by the characteristic peaks of PGNAA spectrum output.
\item Classification with sufficient measure counts is more accurate than with fewer counts; however, the resultant spectrum contains more information than necessary need then the measurement time is too long. If the measurement time is too short, the accuracy of the peaks will be poor, and noise will appear in the spectrum. The minimum value of sample count rate to provide sufficient accuracy will be further discussed in Section \ref{count}.
\item The plottings of the substances having similar compounds are also close to each other because of the chemical similarity. These can be a challenge in this project.
\end{enumerate}

\section{Methods}
In order to construct a well-defined and modularized system architecture, we started our workflow with crucial procedures such as \textbf{designing Random Sampling Methods}, \textbf{constructing models architectures}, \textbf{training} and \textbf{optimizing the training set-up}. During this phase, we also have to foresee some of the future needs such as maintenance, revision, or expansion in the future.
\subsection{Random Sample Methods (RSMs)}\label{sec:rsm}
\begin{figure}[t]
\centerline{\includegraphics[width=2.2in]{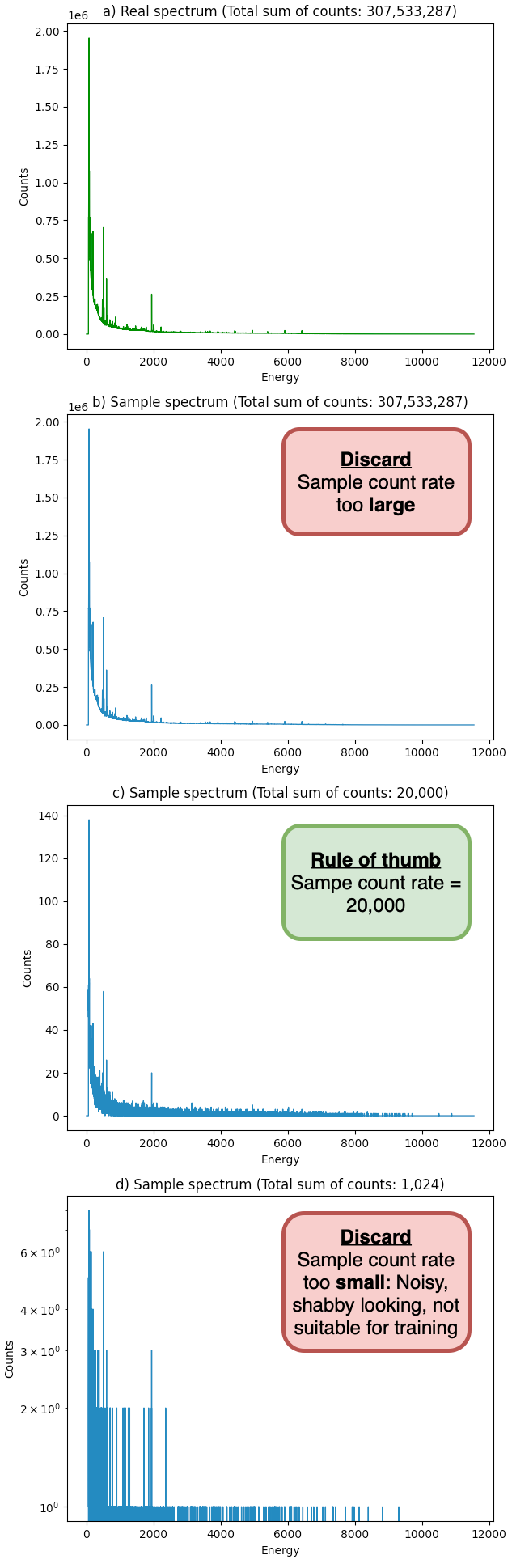}}
\caption{Sampling PGNAA spectrum with different sample count rates. Spectra show clearly visible characteristic peaks for high count rates. Those peaks vanish in noise for low count rates. }
\label{fig1}
\end{figure}
Random Sample Methods (RSMs) is a novel idea that could fulfil our main goal of reducing the measurement time \cite{helmand}, which is corresponded to “counts”. The reason to conceive creating downsized samples with random methods is that their nature resembles the original PGNAA spectrum. Their similar nature help to identify the constituent substances and expedite the classification process. A complete data source can be collected all at once. The resulted sample spectrum is the skimmed version of its origin, and its measuring time is reduced tremendously. In other words, spectrum scanning of the target material takes originally few hours, and only a few seconds is allowed for our sample process. Hence, we generate down-sized samples for every training iteration. By applying RSM on the dataset, one detailed spectrum for each composition material is already enough train a model \cite{dl,cnn}, e.g. a CNN.

Let’s use an imaginative and distinctive energy level list = [1.3 keV, 2.6 keV, 3.9 keV, 5.2 keV]
, then we have an over-simplified spectrum from PGNAA equipment and treat as the full measuring sample data corresponding to a larger measuring time: [3.9 keV, 2.6 keV, 2.6 keV, 2.6 keV, 2.6 keV, 2.6 keV, 2.6 keV, 2.6 keV, 3.9 keV, 2.6 keV, 2.6 keV, 1.3 keV, 2.6 keV].

\begin{itemize}
\item For saving memory space, the data stored into a CSV file (in our data source) as two arrays: the above distinctive energy level list and its corresponding count list = [1, 10, 2, 0] are recorded.
\end{itemize}

For RSM, we randomly choose a smaller sample, corresponding to a shorter measuring time, for examples five samples (k = 5). The output of this sampling method is a downsized sample as a lengthy list of energy levels, and it might look like this: [2.6 keV, 1.3 keV, 2.6 keV, 3.9 keV, 2.6 keV]. Using the above distinctive energy level list = [1.3 keV, 2.6 keV, 3.9 keV, 5.2 keV] and counting the corresponding occurrence, the output of RSM is: 
\begin{itemize}
\item Result of RSM = corresponding count list = [1, 3, 1, 0]
\end{itemize}
RMS can be repeated arbitrarily often, to lead arbitrarily many data points for training machine learning models.
\subsection{CNN design for one-dimensional data}
Convolutional Neural Network (CNN) has been widely used in the field of supervised learning \cite{cnn} and especially with image classification. The filters (weights) are matrix values, which can be considered as feature extractor, are learned during the training phase of the model. A convolution between a filter and an image can induce effects such as sharpness, edge detection, and even the characteristic peaks of our spectrum.

After rounds of testing and comparisons, see Table \ref{tab2} and \cite{benchmark}, the Inception-ResNet-V2 model was usually significantly faster than their competitors and offer a better throughput-accuracy tradeoff. It also has relatively fewer model blocks, thus less work is needed for the modification in comparison with other complex architectures.

The following steps are taken to handle one-dimensional input data so that it can be compatible with our speedy RSM:
\begin{itemize}
\item Replace/ add an untrained convolution layer with custom variants \cite{replace} at the initial point of the pretrained network (“Stem” in Inception model), so that the dimension of input data can be fed.

\item Set desired output classes.

\item Reduce all 2D layers to 1D including the kernel dimension. 
\end{itemize}

\subsection{Training and Fine Tuning}
Besides optimizing the common training hyper-parameters, like batch size, optimizer, learning rate, etc., we also want to minimize our PGNAA data by (1) seeking the lowest count rate (i.e. live time) of PGNAA samples and (2) seeking energy channels that can be discarded for both CNN training and prediction.

\begin{figure}[t]
\centerline{\includegraphics[width=3.8in]{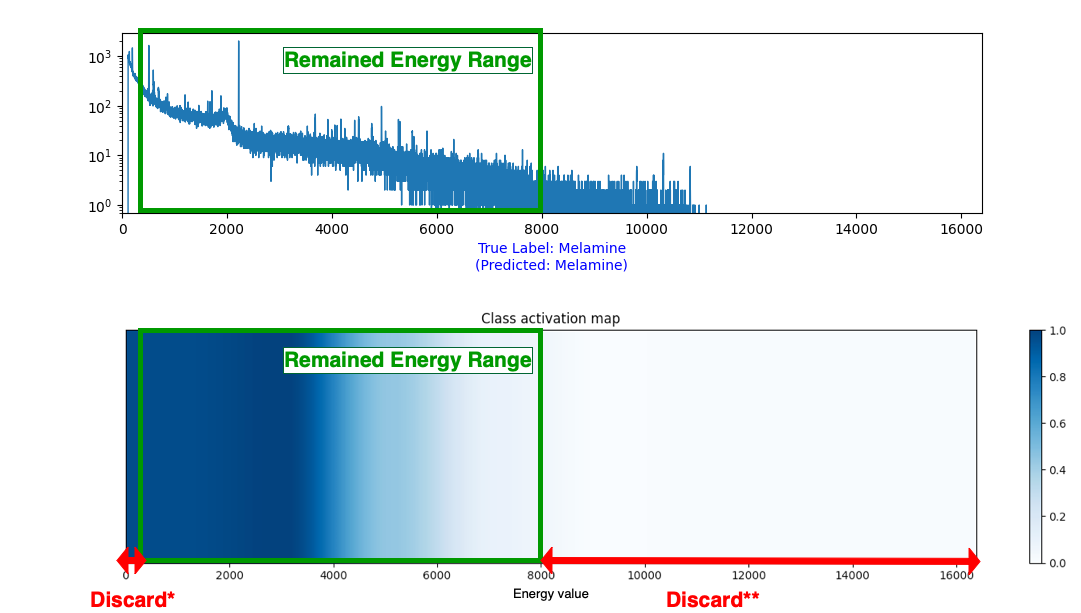}}
\caption{Visualize CNN prediction process with CAM. \textit{Class Activation Map (CAM) of a specific category illustrates how CNN recognizes the distinguished image regions of that category, by using Global Average Pooling (GAP) performed before the last convolutional network layer.}}
\label{fig2}
\end{figure}

\begin{enumerate}
\item The measurement count of a sample corresponds linearly to the measurement time of a sample; the fewer the count of measurements, the shorter the measurement time, and the more difficult it is to classify the sample. In short, the reduction of the sampling time is essential for this project, but excessive reduction of measurement time could lead to very noisy data, see \figurename~\ref{fig1}. 

\item Class Activation Map (CAM) can be used to illustrate the prediction decisions made by CNN, since Class Activation Map of a specific category indicates how CNN recognizes the distinguished image regions of that category \cite{cam}. 

\end{enumerate}

As shown in \figurename~\ref{fig2}, CAM is applied to squeeze the data further in order to expedite the classification process. The program’s original setting for the CAM is two dimensions, the setting was changed to one dimension because our selected CNN Model requires one dimension; for instance, the input and the convolutions are one dimension. Base on the result graph, we discard Channel 8000 to Channel 16384 (5641.92 keV to 11552.48 keV) on the high energy band in the later setting for Experiment I.  

The final decision is 0 to 103 on the lower energy band and C8000 to 16384 on the high energy band should be filtered out so that the accuracy and training time could be optimized.

Summarizing above, the final tuning of the best training for “10 species of substances and two species of soil” is shown as Table \ref{table}.
\begin{table}[htbp]
\caption{Training Set-up for Experiment I}
\label{table}
\setlength{\tabcolsep}{3pt}
\begin{tabular}{|p{95pt}|p{145pt}|}
\hline
Random Sampling Method & Random weighted selection in counts\\
CNN Model & CNN Inception-Resnet-V2 modified 1D\\
Batch size & 128 (hardware limit)\\
Optimizer & Adam \cite{adam}\\
Learning rate \cite{lr} & 0.01\\
Epoch & 150 (stop when the wish accuracy is reached)\\
Discarded energy range & 0 - 103 and 8000 - 16348 \\
Sample count rate & 19650 (live time $\approx 2.479 s$)\\
\hline
\end{tabular}
\label{tab1}
\end{table}
\section{Experimental results}\label{sec:results}

\subsection{Experiment I: Train CNN with Dataset of 10 species of substances and two species of soil (live time $\approx$ 2.479 sec)}
12 CSV files are used for training purposes which consists of 10 species of sample substances and two species of soil. These files fed into the RSM to be downsized and “smashed”. Totally 19,000 samples are generated and fed into the CNN model for training which requires 150 epochs of iterations. 

As shown in Table \ref{cm-exp1}, the trained CNN model makes good prediction and achieve good results; the accuracy is averagely about 96.88\%. The prediction mostly matches with the true label. However, some misclassifications occurred ($\approx$ 68.5\%) that occurred between “Erdreich-11-15-30” and “Erdreich-HgS-inhomogen”. This “same species” classification problem will be handled in the next experiment.

\subsection{Experiment II: Dataset Aluminium with 20,000 as sample count rate (live time $\approx$ 1.11 sec)}
Using the same set-up as Experiment I, such as the same CNN model, same RSM, only the input dataset is using “Dataset Aluminium” and the discarded energy band of “0 - 103” for this new training. The result is schown in Table \ref{cm-exp2}

Even when the training is only focused on the substances with the same species, the performance remains about 65 \% and seems impossible to be raised. The problem can be caused by the setting of the sample count rate, in which the resulted sample data from RSM is too noisy and doesn’t provide enough information to distinguish from similar species.

\subsection{Experiment IIIa: Dataset Aluminium with increased sample count rate to 500,000 (live time $\approx$ 27.74 s)}\label{count}

\begin{figure}[t]
\centerline{\includegraphics[width=3.8in]{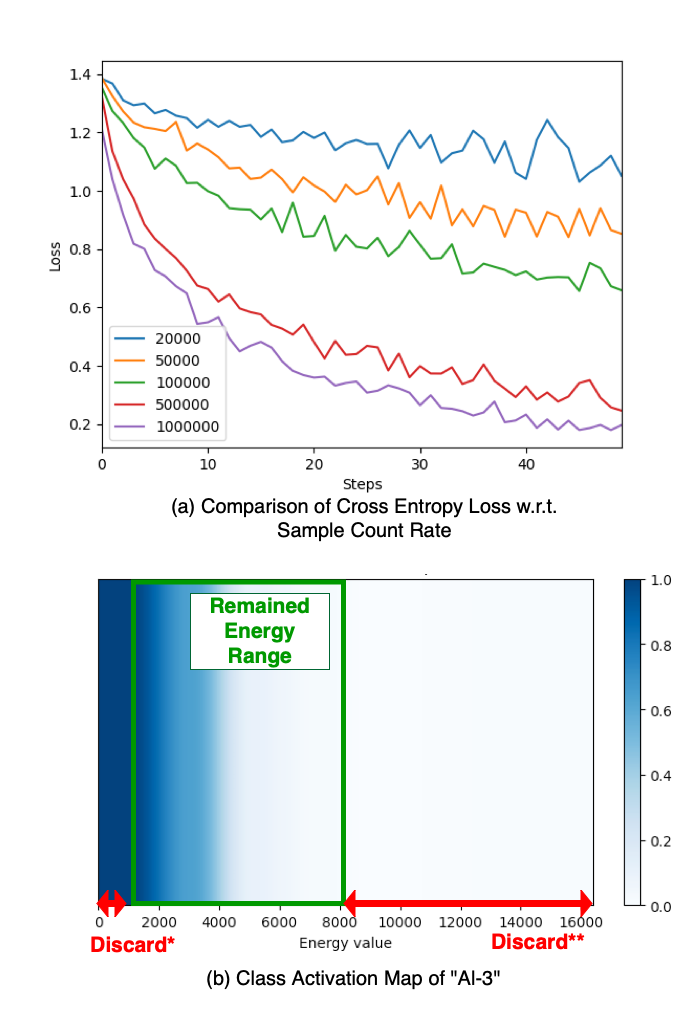}}
\caption{Optimizing sampling strategy for Dataset Aluminium. \textit{Cross Entropy Loss between input and target is computed during training a classification problem. The aim of training is to minimize the loss.}} 
\label{fig3}
\end{figure}

Totally five different sample count rates were tested. The curve with the lowest loss function and the best convergence is chosen. We started with the count rate (20,000) suggested by the last experiment. At the end, we choose 500,000 as our sample count rate according \figurename~\ref{fig3}a, in which the loss function descents stably. The result is shown in Table \ref{cm-exp3a}.

In addition, we visualized in \figurename~\ref{fig3}b with CAM which energy ranges have more influence on classification and filtered out the irrelevant energy range to save training time.

\subsection{Experiment IIIb: Dataset Copper with increased sample count rate to 500,000 (live time $\approx$ 24.44 s)}
For Dataset from Wieland’s Copper, we repeated the same process as Experiment IIIa to choose the optimal sample count rate for training. The only difference is the CNN’s high energy discarded range is set to the range of 14,000 to 16,384 (base on the suggestion from CAM) in order to improve the training speed. The result is shown in Table \ref{cm-exp3b}

\section{Discussion}
All of our designed models, which equips with tested and chosen RSMs, can classify the PGNAA spectrum and find out the constituent materials speedily. The CNNs can classify the sampled PGNAA spectrum based on automated extracted aggregate features, which is different from common Machine Learning methods like Support Vector Machine (SVM) \cite{svm}. Other classical approaches such as Normalized Cross Correlation (NCC) \cite{ncc} and Chi-squared matching, perform the classification by comparing the input data with templates iteratively. In order to achieve better accuracy, the number of these templates must be increased since they are proportionally related.

\section{Conclusion}
The essence of this essay is to devise a suitable Random Sampling Method (RSM) to generate some downsized samples to train the CNN models instead of the cumbersome fully measured samples. The downsized samples are proven to be easier to handle and improving speed. Also, the downsized samples have a resemblance nature to the test spectrum reading from the PGNAA because both are heavily downsized data. These two reasons boost PGNAA’s reading speed so fast that it becomes a high-speed scanner.

In the end, reduced Model (CNN Inception-ResNet-V2modified 1D) using RSM (Random weighted selection in counts) to extract our data from CSV, was determined as our best matching pair after several test runs.

Besides tuning the model with different batch sizes, optimizers, and learning rates, the downsized training samples were also inspected with different sampling count rates so as to minimize our necessary sample live time for a prediction. In addition, to strengthen the downsizing process, some less important energy channels were filtered out according to the energy information provided by the Class Activation Map. For example, the highest energy range (Channel 8000 - 16384) and the lowest energy range are discarded.

\appendices
\section*{Data Availability}
The MetalClas project is funded by the German Federal Ministry of Education and Research (BMBF) with grant number 01IS20082B. The copper alloys were provided by Wieland-Werke AG. The PGNAA data used to support the findings of this study are available in a public repository for AI Challenge Days 2021.

\section*{Appendix}
\subsection{Comparing RSM and their best models}
Four RSM were designed, but in section \ref{sec:rsm} only RSM-3 was introduced. The other three RSM are implemented using Python as follow: 
\begin{itemize}
\item RSM-1 is simply the lengthy measured energy value list. 

\item RSM-2 applies a random binomial function to each row in the column ’Counts’ of the CSV files. Nevertheless, the sum of measurement counts within a sample remains similar for each material.

\item RSM-4a or RSM-4b is a scatter-plot format or a histogram format converted from the sampling result from RSM-1. 
\end{itemize}
The four RSM methods have to test together with DL models. In this section, CNN and other DL models were tested together with RSM. 
\begin{table}[H]
\caption{Comparing RSM and their best models}

\setlength{\tabcolsep}{3pt}
\begin{tabular}{|c|p{125pt}|c|c|c|}
\hline
& Combination of model and RSM & Accuracy & Sampling & Iteration\\
&&& time (s) & time (s)\\
\hline
1 & CNN ResNet-101 + RSM-1 & $\approx$ 65\% & 0.012 & -\\
2 & CNN custom 1D + RSM-3 & $\approx$ 65\% & 0.073 & -\\
3 & \scriptsize{CNN ResNet-101 (modified 1D) + RSM-2} & - & 1.488 & -\\
4 & \scriptsize{CNN ResNet-101 (modified 1D) + RSM-3} & $\approx$ 80\% & 0.073 & - \\
5 & \scriptsize{CNN Inception-ResNet-V2 \cite{incp}+ RSM-2} & - & 1.488 & - \\
\textbf{6} & \textbf{CNN Inception-ResNet-V2 + RSM-3} & $\approx$ 90\% & 0.073 & 2.1 \\
7 & CNN NasnetALarge \cite{nasnet} + RSM-4a & $<$ 80\% & 0.696 & -\\
8 & CNN PNasnet5Large \cite{pnas}+ RSM-4a & $<$ 80\% & 0.696 & -\\
9 & RNN \cite{transformer} custom + RSM-3 & $<$ 40\% & 0.073 & - \\
10 & Transformer \cite{transformer} (custom) + RSM-3 & $\approx$ 90\% & 0.073 &13.8 \\
\hline
\end{tabular}
\label{tab2}
\end{table}

\subsection{Confusion Matrices for Experiment I, II, III$a$ and III$b$}
\newcommand\items{12} 
\arrayrulecolor{white} 
\noindent
\begin{table}[H]
\caption{Confusion Matrix for Experiment I}\label{cm-exp1}
\centering
\resizebox{9cm}{!}{
\begin{tabular}{cc*{\items}{|E}|}

\multicolumn{1}{c}{} &\multicolumn{1}{c}{} &\multicolumn{\items}{c}{Predicted} \\ \hhline{~*\items{|-}|}
\multicolumn{1}{c}{} & 
\multicolumn{1}{c}{} & 
\multicolumn{1}{c}{\rot{Scrap metal powder}} & 
\multicolumn{1}{c}{\rot{Cement}} & 
\multicolumn{1}{c}{\rot{Stucco}} &
\multicolumn{1}{c}{\rot{Al-1}} &
\multicolumn{1}{c}{\rot{Cu-1}} &
\multicolumn{1}{c}{\rot{Melamine}} &
\multicolumn{1}{c}{\rot{Asilikos}} &
\multicolumn{1}{c}{\rot{PVC}} & 
\multicolumn{1}{c}{\rot{Soil-1}} &
\multicolumn{1}{c}{\rot{Soil Hgs inhomogeneous}} &
\multicolumn{1}{c}{\rot{Battery NiCd}} &
\multicolumn{1}{c}{\rot{Copper ore-A-prod}} \\ 
\hhline{~*\items{|-}|}
\multirow{\items}{*}{\rotatebox{90}{Actual}} 
&Scrap metal powder & 1 &  0  &  0  &  0  &  0  &  0  &  0  &  0  & 0  &  0  &  0  &  0  \\ \hhline{~*\items{|-}|}
&Cement  & 0 &  1  &  0  &  0  &  0  &  0  &  0  &  0  & 0  &  0  &  0  &  0   \\ \hhline{~*\items{|-}|}
&Stucco  & 0 &  0  &  1  &  0  &  0  &  0  &  0  &  0  & 0  &  0  &  0  &  0  \\ \hhline{~*\items{|-}|}
&Al-AW7075  & 0 &  0  &  0  &  1  &  0  &  0  &  0  &  0  & 0  &  0  &  0  &  0  \\ \hhline{~*\items{|-}|}
&Cu  & 0 &  0  &  0  &  0  &  1  &  0  &  0  &  0  & 0  &  0  &  0  &  0  \\ \hhline{~*\items{|-}|}
&Melamine  & 0 &  0  &  0  &  0  &  0  &  1  &  0  &  0  & 0  &  0  &  0  &  0  \\ \hhline{~*\items{|-}|}
&Asilikos  & 0 &  0  &  0  &  0  &  0  &  0  &  1  &  0  & 0  &  0  &  0  &  0  \\ \hhline{~*\items{|-}|}
&PVC  & 0 &  0  &  0  &  0  &  0  &  0  &  0  &  1 & 0  &  0  &  0  &  0  \\ \hhline{~*\items{|-}|}
&Soil-1  & 0 &  0  &  0  &  0  &  0  &  0  &  0  &  0  & 1  &  0  &  0  &  0  \\ \hhline{~*\items{|-}|}
&Soil Hgs inhomogeneous  & 0 &  0  &  0  &  0  &  0  &  0  &  0  &  0  & 0.28  &  0.72  &  0  &  0  \\ \hhline{~*\items{|-}|}
&Battery NiCd  & 0 &  0  &  0  &  0  &  0  &  0  &  0  &  0  & 0  &  0  &  1  &  0  \\ \hhline{~*\items{|-}|}
&Copper ore-A-prod  & 0 &  0  &  0  &  0  &  0  &  0  &  0  &  0  & 0  &  0  &  0  &  1  \\ \hhline{~*\items{|-}|}
\end{tabular}
}
\end{table}

\newcommand\fouritems{4}   
\noindent
\begin{table}[H]
\caption{Confusion Matrix for Experiment II}\label{cm-exp2}
\centering
\resizebox{4.8cm}{!}{
\begin{tabular}{cc*{\fouritems}{|E}|}
\multicolumn{1}{c}{} &\multicolumn{1}{c}{} &\multicolumn{\fouritems}{c}{Predicted} \\ \hhline{~*\fouritems{|-}|}
\multicolumn{1}{c}{} & 
\multicolumn{1}{c}{} & 
\multicolumn{1}{c}{\rot{Al-1}} & 
\multicolumn{1}{c}{\rot{Al-2}} & 
\multicolumn{1}{c}{\rot{Al-3}} &
\multicolumn{1}{c}{\rot{Al-4}} \\ 
\hhline{~*\fouritems{|-}|}
\multirow{\fouritems}{*}{\rotatebox{90}{Actual}} 
&Al-1 & 0.97 &  0  &  0  &  0.03 \\ \hhline{~*\fouritems{|-}|}
&Al-2  & 0.06 &  0.38  &  0.41  &  0.16    \\ \hhline{~*\fouritems{|-}|}
&Al-3  & 0 &  0.12  &  0.88  &  0   \\ \hhline{~*\fouritems{|-}|}
&Al-4  & 0.53 &  0.22  &  0.03  &  0.22 \\ \hhline{~*\fouritems{|-}|}
\end{tabular}
}
\end{table}

\noindent
\begin{table}[H]
\caption{Confusion Matrix for Experiment III$a$}\label{cm-exp3a}
\centering
\resizebox{4.8cm}{!}{
\begin{tabular}{cc*{\fouritems}{|E}|}
\multicolumn{1}{c}{} &\multicolumn{1}{c}{} &\multicolumn{\fouritems}{c}{Predicted} \\ \hhline{~*\fouritems{|-}|}
\multicolumn{1}{c}{} & 
\multicolumn{1}{c}{} & 
\multicolumn{1}{c}{\rot{Al-1}} & 
\multicolumn{1}{c}{\rot{Al-2}} & 
\multicolumn{1}{c}{\rot{Al-3}} &
\multicolumn{1}{c}{\rot{Al-4}} \\ 
\hhline{~*\fouritems{|-}|}
\multirow{\fouritems}{*}{\rotatebox{90}{Actual}} 
&Al-1 & 0.98 &  0  &  0  &  0.02   \\ \hhline{~*\fouritems{|-}|}
&Al-2  & 0 &  1  &  0 &  0    \\ \hhline{~*\fouritems{|-}|}
&Al-3  & 0 &  0.1  &  0.9  &  0   \\ \hhline{~*\fouritems{|-}|}
&Al-4  & 0.06 &  0.09  &  0  &  0.84   \\ \hhline{~*\fouritems{|-}|}
\end{tabular}
}
\end{table}

\newcommand\fiveitems{5}   
\noindent
\begin{table}[H]
\caption{Confusion Matrix for Experiment III$b$}\label{cm-exp3b}
\centering
\resizebox{6.0cm}{!}{
\begin{tabular}{cc*{\fiveitems}{|E}|}
\multicolumn{1}{c}{} &\multicolumn{1}{c}{} &\multicolumn{\fiveitems}{c}{Predicted} \\ \hhline{~*\fiveitems{|-}|}
\multicolumn{1}{c}{} & 
\multicolumn{1}{c}{} & 
\multicolumn{1}{c}{\rot{Cu-1}} & 
\multicolumn{1}{c}{\rot{Cu-2}} & 
\multicolumn{1}{c}{\rot{Cu-3}} &
\multicolumn{1}{c}{\rot{Cu-4}} &
\multicolumn{1}{c}{\rot{Cu-5}}\\ 
\hhline{~*\fiveitems{|-}|}
\multirow{\fiveitems}{*}{\rotatebox{90}{Actual}} 
&Cu-1 & 0.97 &  0.03  &  0  &  0 & 0   \\ \hhline{~*\fiveitems{|-}|}
&Cu-2  & 0.09 &  0.91  &  0  &  0 & 0    \\ \hhline{~*\fiveitems{|-}|}
&Cu-3 & 0 &  0 & 1  &  0 &  0   \\ \hhline{~*\fiveitems{|-}|}
&Cu-4 & 0 &  0  &  0  & 1 & 0 \\ \hhline{~*\fiveitems{|-}|}
&Cu-5 & 0 &  0  &  0  &  0 & 1   \\ \hhline{~*\fiveitems{|-}|}
\end{tabular}
}
\end{table}


\begin{thebibliography}{00}

\bibitem{pgnaa1} J.-L Ma, C. Carasco, Bertrand Perot, E Mauerhofer, John Kettler, and Andreas Havenith. “Prompt gamma neutron activation analysis of toxic elements in radioactive waste packages”. In: Applied radiation and isotopes : including data, instrumentation and methods for use in agriculture, industry and medicine 70 (Feb. 2012), pp. 1261-3. doi: 10.1016/j.apradiso.2012.02.011

\bibitem{pgnaa2} H. P. Chang, I. Meric, D. Sudac, K. Nad, J. Obhodas, G. Hou, Y. Zhang, and R. P. Gardner. “Implementation of the Monte Carlo Library Least-Squares (MCLLS) approach for quantification of the chlorine impurity in an online crude oil monitoring system”. In: Radiation Physics and Chemistry 155 (2019). IRRMA-10, pp. 197-201. issn: 0969-806X. doi: https://doi.org/10.1016/j.radphyschem.2018.05.012. url: https://www.sciencedirect.com/science/article/pii/S0969806X17308095

\bibitem{pgnaa3} H. Shahabinejad, N. Vosoughi, and F. Saheli. “Processing oscintillation gamma-ray spectra by artificial neural network”. In: Journal of Radioanalytical and Nuclear Chemistry 325.2 (Aug. 2020), pp. 471-483. issn: 1588-2780. doi: 10.1007/ s10967-020-07239-w. url: https://doi.org/10.1007/s10967-020-07239-w

\bibitem{helmand} H. Shayan, K. Krycki, M. Doemeland, M. Lange-Hegermann. “PGNAA Spectral Classification of Metal with Density Estimations”. In: 23rd Virtual IEEE Real Time Conference (Aug. 2022)

\bibitem{dash} D. Muthukrishna, D. Parkinson, and B. Tucker. “DASH: Deep Learning for the Automated Spectral Classification of Supernovae and Their Hosts”. In: The Astro- physical Journal 885 (Nov. 2019), p. 85. doi: 10.3847/1538-4357/ab48f4 (pages 3, 27)

\bibitem{noise} R. E. Hendrick. “Sampling time effects on signal-to-noise and contrast-to-noise ratios in spin-echo MRI. Magn Reson Imaging”. 1987;5(1):31-7. doi: 10.1016/0730-725x(87)90481-4. PMID: 3586870.

\bibitem{dl} S. Saxena. “Artificial Neuron Networks(Basics) | Introduction to Neural Networks”. [Online; accessed 5-May-2021]. 2017. url: https://becominghuman.ai/artificial-neuron-networks-basics-introduction-to-neural-networks-3082f1dcca8c (page 11).

\bibitem{cnn} C. Szegedy, W. Liu, Y. Jia, P. Sermanet, S. E. Reed, D. Anguelov, D. Erhan, V. Vanhoucke, and A. Rabinovich. “Going Deeper with Convolutions”. In: CoRR abs/1409.4842 (2014). arXiv: 1409.4842. url: http://arxiv.org/abs/1409.4842

\bibitem{benchmark} S. Bianco, R. Cadène, L.Celona, and P. Napoletano. “Benchmark Analysis of Rep- resentative Deep Neural Network Architectures”. In: CoRR abs/1810.00736 (2018). arXiv: 1810.00736. url: http://arxiv.org/abs/1810.00736 (page 18).

\bibitem{replace} J. Brownlee “How Do Convolutional Layers Work in Deep Learning Neural Networks?” [Online; accessed 20-Nov-2021]. url: https://machinelearningmastery.com/convolutional-layers-for-deep-learning-neural-networks/

\bibitem{cam} B. Zhou, A. Khosla, Lapedriza. A., A. Oliva, and A. Torralba. “Learning Deep Features for Discriminative Localization”. In: CVPR (2016)

\bibitem{adam} D. P. Kingma und J. Ba “Adam: A Method for Stochastic Optimization” 2015

\bibitem{lr} J. Patterson und A. Gibson “Deep learning: A practitioner’s approach”, First Hrsg., Sebastopol CA: O’Reilly, 2017

\bibitem{svm} F. Xing, P. Guo “Classification of Stellar Spectral Data Using SVM”. In: Yin FL., Wang J., Guo C. (eds) Advances in Neural Networks - ISNN 2004. Lecture Notes in Computer Science, vol 3173. Springer, Berlin, Heidelberg.

\bibitem{ncc} F. Zhao, Q. Huang, and W. Gao. “Image Matching by Normalized Cross-Correlation”. In: vol. 2. June 2006, pp. II-II. doi: 10.1109/ICASSP.2006.1660446

\bibitem{incp}V. Vanhoucke C. Szegedy S. Loffe. “Inception-v4, Inception-ResNet and the Impact of Residual Connections on Learning”. In: CoRR abs/1602.07261 (2016). arXiv: 1602.07261. url: http://arxiv.org/abs/1602.07261 (page 19).

\bibitem{nasnet} B. Zoph and Q. V. Le. “Neural Architecture Search with Reinforcement Learning”. In: CoRR abs/1611.01578 (2016). arXiv: 1611.01578. url: http://arxiv.org/ abs/1611.01578 (page 20).

\bibitem{pnas} C. Liu et al. “Progressive Neural Architecture Search”. In: BMC Bioinformatics 8 (2018). (page 20).

\bibitem{transformer} A. Vaswani, N. Shazeer, N. Parmar, J. Uszkoreit, L. Jones, A. N. Gomez, L. Kaiser, and I. Polosukhin. “Attention is All you Need”. In: Advances in Neural Information Processing Systems. Ed. by I. Guyon, U. V. Luxburg, S. Bengio, H. Wallach, R. Fergus, S. Vishwanathan, and R. Garnett. Vol. 30. Curran Associates, Inc., 2017. (page 22)


\end{thebibliography}
\end{document}